\documentclass[12pt,a4paper,twoside, twocolumn]{article}
\usepackage{dhasa}
\usepackage{titlesec}
\usepackage{algorithm, algpseudocode}

\usepackage{amsmath,scalerel}
\usepackage{graphicx}
\usepackage{tabularx}
\usepackage{soul}
\usepackage{epstopdf}
\usepackage[utf8]{inputenc}

\begin{document}

\title*{Textual Augmentation Techniques Applied to Low Resource Machine Translation: Case of Swahili}


\textit{Gitau, Catherine}\\
\textit{African Institute for Mathematical Sciences Ghana} \\
\textit{cgitau@aimsammi.org} 

\textit{Marivate, Vukosi}\\
\textit{Department of Computer Science, University of Pretoria, South Africa} \\
\textit{vukosi.marivate@cs.up.ac.za} 

\section*{Abstract}
In this work we investigate the impact of applying textual data augmentation tasks to low resource machine translation. There has been recent interest in investigating approaches for training systems for languages with limited resources and one popular approach is the use of data augmentation techniques. Data augmentation aims to increase the quantity of data that is available to train the system. In machine translation, majority of the language pairs around the world are considered low resource because they have little parallel data available and the quality of neural machine translation (NMT) systems depend a lot on the availability of sizable parallel corpora. We study and apply three simple data augmentation techniques popularly used in text classification tasks; synonym replacement, random insertion and contextual data augmentation and compare their performance with baseline neural machine translation for English-Swahili (En-Sw) datasets. We also present results in BLEU, ChrF and Meteor scores. Overall, the contextual data augmentation technique shows some improvements both in the $EN \rightarrow SW$ and $SW \rightarrow EN$ directions. We see that there is potential to use these methods in neural machine translation when more extensive experiments are done with diverse datasets.

Keywords: low-resource, data augmentation, machine translation

\section{Introduction}

There have been several advancements in machine translation and modern MT systems that can achieve near human-level translation performance on the language pairs that have significantly large parallel training resources. Unfortunately, neural machine translation systems perform very poorly on low-resource language pairs where parallel training data is scarce. Improving translation performance on low-resource language pairs could be very impactful considering that these languages are spoken by a large fraction of the world's population.

According to \cite{encyclopedia}, only about 5-10 million people speak Swahili as their native language, but it is spoken as a second language by around 80 million people in Southeast Africa lingua franca, making it the most widely spoken language of sub-Saharan Africa.

Despite the fact that the language is spoken by millions across the African continent, it accounts for less than 0.1\% of the internet whereas 58.4\% of the Internet's content is in English according to \cite{W3Techs}, making it a low-resourced language. Even though Swahili is spoken by so many people, there is little extensive work that has been done to improve translation models built for the language. Data that is needed to produce high quality neural machine translation systems is unavailable resulting in poor translation quality. 

In computer vision, data augmentation techniques are used widely to increase the robustness and improve the learning of the objects with very little training examples. In image processing, the trained data is augmented by, for example, horizontally flipping the images, random cropping, tilting etc. Data augmentation has now become a standard technique to train deep neural networks for image processing and it is not very common practice in training networks for natural language processing (NLP) tasks such as machine translation. Applying data augmentation techniques in text is not as straightforward as in computer vision because in computer vision, the label and content of the original image is preserved as for natural language processing (NLP) tasks, there is need to retain the context of the sentence after augmentation. There are however several data augmentation methods that have been proven to improve performance on various NLP tasks such as text classification but it is not common practice to apply these data augmentation methods for tasks such as machine translation.

Neural machine translation (NMT) as presented in the work of \cite{DBLP:journals/corr/SutskeverVL14};\cite{cho-etal-2014-properties} is a sequence to sequence task that uses a bidirectional recurrent neural network known as an encoder to process a source sentence into vectors called the decoder which then predicts words in the target language. To be able to train a model that is able to produce good translations, these networks require a lot of parallel data or sentence pairs with words that are occurring in diverse contexts which is not available in low-resource language pairs therefore making the performance of the models quite low. One of the solutions to this problem is to manually annotate the available monolingual data which is time consuming and expensive or to perform unsupervised data augmentation techniques. 

In this work we explore some data augmentation techniques that are widely used to improve text classification tasks and investigate their impact on the performance on low resource neural machine translation models for English-Swahili(En-Sw). The most popular text augmentation techniques applied to text classification tasks consist of four powerful operations; synonym replacement, random insertion, random swap and random deletion. These methods have been shown to improve performance in text classification tasks as shown in \cite{wei-zou-2019-eda}. And to better gauge the effect of our data augmentation methods, we compare the results with a baseline model trained on En-Sw datasets.

In summary, our contributions include:
\begin{enumerate}
    \item We explore and evaluate on NMT task, three data augmentation techniques currently only being used in text classification tasks; Word2Vec based-augmentation which does synonym replacement in the sentences, TF-IDF-based augmentation to insert words in random positions in the sentence as well as use of Masked Language Model based-augmentation that does contextual data augmentation on the text.
    \item We show how these data augmentation techniques can be used in NMT tasks.
    \item We also extended the textaugment library \cite{Marivate_2020} to use Fasttext's pre-trained models.
    \item We present baseline NMT results in BLEU, Meteor and ChrF scores.
    
\end{enumerate}

This paper is organised as follows; We look at the work that is been done by past authors on data augmentation for low-resource languages first. We also look at the data augmentation techniques and the approaches we used in our study which is described in Section \ref{methodology}. Section \ref{experiments} describes the model settings that were considered for every data augmentation approach, Section \ref{results} discusses the experimental results and then we conclude with stating limitations and future work as well as conclude at Sections \ref{limitations} and \ref{conclusion}.


\section{ Literature Review}

Work on machine translation to improve machine translation quality on low resource languages is a widely studied problem. In natural language processing (NLP), data augmentation is a popular technique that is used to increase the size of the training data.

One promising approach is the use of transfer learning \cite{zoph-etal-2016-transfer}. This method proved that having prior knowledge in translation of a different higher-resource language pair can improve translating a low-resource language. A NMT model is first trained on a large parallel corpus to create the \textit{parent model} and continued to train this model by feeding it with a considerable smaller parallel corpus of a low-resource language resulting into the \textit{child model} which inherits the knowledge from the parent model by reusing its parameters. The parent and child language pairs shared the same target language(English). The use of data from another language can be seen as a data augmentation task in itself and large improvements have been observed especially when the high-resource language is syntactically similar with the low-resource language \cite{lin-etal-2019-choosing}.

The work of \cite{sennrich-etal-2016-neural} explores a data augmentation method for machine translation known as back translation where machine translation is used to automatically translate target language monolingual data into source language data to create synthetic parallel data for training and is currently the most commonly used data augmentation technique in machine translation tasks. The quality of the backward system while effective, has been shown to negatively affect the performance of the final NMT Model when the target-side monolingual data is limited. Back translation as a method for performing data augmentation in machine translation could deteriorate the Low resource Language - English(LRL - ENG) translation performance due to the limited size of the training data as shown in \cite{xia-etal-2019-generalized}.

\cite{xia-etal-2019-generalized} augment parallel data through two methods: back-translating from ENG to low resource language (LRL) or high resource language (HRL) and converting the HRL-ENG dataset to a pseudo LRL-ENG  dataset. They use an induced bilingual dictionary to inject LRL words into the HRL then further modify these sentences using modified unsupervised machine translation framework. Their method proved to improve translation quality as compared to supervised back-translation baselines however, the method requires access to a HRL that is related to the LRL as well as monolingual LRL.

There are other data augmentation methods which have been used in other NLP tasks such as text classification to improve performance. \cite{wei-zou-2019-eda} show that simple word replacement using knowledge bases like WordNet \cite{Miller95wordnet:a} can improve performance of classification tasks. \cite{Marivate_2020} also observe that Word2Vec-based augmentation is also a viable solution when one does not have access to a knowledge base of synonyms such as the WordNet \cite{Miller95wordnet:a}. \cite{kumar-etal-2020-data} show that seq2seq pre-trained models can be effectively used for data augmentation and these provide the best performance. These data augmentation methods are currently only being used to improve classification tasks and have not yet been utilized in any neural machine translation task to improve performance. In this work we will be looking at how some of these methods can be used to also improve neural machine translation models where the data is low-resourced. In particular, we will explore three data augmentation methods which include: 1) Word2Vec based-augmentation, 2) Tf-idf based augmentation, 3) Masked Language Model based-augmentation and use the additional data to train the NMT model.

\section{Methodology}
\label{methodology}
Our goal is to compare different data augmentation methods that are used in text classification tasks with the aim of identifying whether the methods can be used to improve the baseline NMT score. The results are compared across two different datasets and uses in-domain test sets to demonstrate the generalization capability of the models. These experiments are useful to help other researchers gain insights as they work on building better neural machine translation models for low-resourced languages. First, we describe the data that was used to train the models, then the data augmentation methods that we will be using and finally give details of the experiments we performed to test these methods together with the results obtained. 

\subsection{Training Data}
Small amounts of parallel data are available for Swahili-English. The data was received from the work of \cite{2020arXiv200314402L} where they released standardized experimental data and test sets for five different languages(Swahili, Amharic, Tigirinya, Oromo and Somali). They collected all available parallel corpora for those five languages from the Opus corpus \cite{tiedemann-2012-parallel} which consists of a collection of translated texts from the web. For this work, we utilized data that includes JW300 \cite{agic-vulic-2019-jw300} and Tanzil \cite{tiedemann-2012-parallel} which provides a collection of Quran translations to compare with the baseline results from the work of \cite{2020arXiv200314402L}.

Table 1 shows the amount of parallel data that was  collected. The data was split into train, dev and test sets as in \cite{2020arXiv200314402L}. We then segmented the data into subword units using Byte Pair Encoding \cite{sennrich-etal-2016-neural} where we learned $20K$ byte pair encoding tokens.

\begin{table*}[ht]
\begin{center}
\begin{tabular}{|c| c c c c |c|}
    \hline
    \multicolumn{4}{c|}{Domain} \\
    Language Pair & Split  & JW300 & Tanzil & Total  \\
    \hline\hline
      & train & 907842 & 87645 & 1024717\\
    Sw-En & dev & 5179 & 3505  & 8684\\ 
    & test & 5315 & 3509 & 8824\\
    \hline

\end{tabular}
\caption{Data statistics showing number of examples/sentences available across four domains}
\label{tab:table1}
\end{center}
\end{table*}

\subsection{Baseline}

In this approach the Transformer NMT model is trained using Jw300 and Tanzil data combined then tested on different datasets from two different domains (Jw300 and Tanzil). The model is trained with no modifications throughout with standard preprocessing steps such as tokenization, lowercasing and cleaning. This model in this approach serves as a baseline for comparison.

\subsection{Data Augmentation methods}

We augmented the data using three types of augmentation methods: Word2Vec-based augmentation(synonym replacement), Tf-idf based augmentation(random insertion) and Masked Language Model(MLM)-based augmentation(context based augmentation). We combined the first two augmentation methods and used the Masked Language Model-based augmentation on its own. The Word2Vec and Tf-idf augmentations were done on the source language such that when training an En-Sw model, we augment the English language and when training a Sw-En model we augment the Swahili language. In Masked language modeling the augmentation was only done on the English language. 

\subsubsection{Word2vec-based augmentation}
Word2vec is an augmentation technique mostly used in classification tasks that uses a word embedding model \cite{NIPS2013_9aa42b31} that is trained on publicly available datasets to find the most similar words for a given input word. We use Word Vectors pre-trained on Common Crawl and Wikipedia on both English and Swahili data using fastText \cite{joulin-etal-2017-bag} a library for text representation and classification. We load the pre-trained fastText model for each language into our algorithm to augment the texts by randomly selecting a word in the text to determine their similar words using cosine similarity as a relative weight to select a similar word that replaces the input word as done in \cite{Marivate_2020}. Our algorithm is as illustrated in Algorithm \ref{tab:algorithm1} . It receives a string which is the input data and augments the text into five different augmented texts then we use cosine similarity to select the best sentence that is at least 0.85 closer to the original text. The reason for this is that we'd like to retain the contextual meaning of a sentence even after augmentation. We compare the five different augmented sentences and pick the sentence that has a cosine similarity score that is highest. To prevent duplicated augmentations, we drop the sentences that are 100\% similar to the original sentence. This augmentation was done on the source language where the corresponding target language sentences remained constant and unchanged. 
Examples of the augmented sentences can be seen in Table 2.

\begin{table*}[ht]
\begin{center}
 \begin{tabular}{|c||c|} 
 \hline
 Method & Sentence \\ 
 \hline\hline
 \textbf{English} & \\
 Original & The quick brown fox \textbf{jumps} past the lazy dog \\ 
 Word2Vec + tfidf &  The quick brown fox \textbf{leaps} \textbf{over} \textbf{retrorsum} the lazy dog \\
 \hline
 \textbf{Swahili} & \\
 Original & Baba na mama yako ni wazuri sana  \\ 
 Word2Vec + tfidf &\textbf{Kizee} baba na mama yako ni \textbf{wema} \textbf{waar} sana  \\
 [1ex] 
 \hline
 \end{tabular}
 \caption{Table showing example of augmented sentences}
 \label{tab:table2}
 \end{center}
\end{table*}

\subsubsection{Tf-idf based augmentation}
We created another set of augmented data that uses Tf-idf \cite{Ramos2003UsingTT}. The concept of Tf-idf is that high frequency words may not be able to provide much information gain in the text. It means that rare words contribute more weights to the model. In this case, words that have low Tf-idf scores are said to be uninformative and thus can be replaced or inserted in text without affecting the ground truth labels of the sentence. Here, the words that are chosen to be inserted at a random position in the sentence are chosen by calculating the Tf-idf scores of words over all the sentences and then taking the lowest ones. We therefore insert a new word at a random position according to the Tf-idf calculation. This was also done on the source language only and the corresponding target language sentences remain unchanged. Tf-idf was applied after performing the Word2Vec based augmentation method. This is illustrated in Algorithm \ref{tab:algorithm1}.

\begin{algorithm}
  \label{algorithm1}
  \begin{algorithmic}[1]
  \Require{$s$: a sentence } 
  \Ensure{$\hat{s}$: augmented sentence}
  \\ \textit{Step 1: get similar words of each word in $s$} :
  \Procedure{Augment}{$s$}\Comment{Augmentation of sentence}
    \State $t \gets$ sentence s tokenized 
    \State $u \gets$ unique words from $t$
    \For{$w$ \textit{in}$(u)$}
    \State $\overrightarrow{w} \gets$ find five similar words for $w$ 
    \EndFor
    \\ \textit{Step 2: replace random words in $s$ with similar words and insert Tf-idf word} :
    \State $n \gets$ 5
    \For{$\_$ \textit{in range}$(n)$}
        \State $w_i\gets$ randomly select a word from $s$
        \State $w_0\gets$ randomly select one similar word for $w_i$ from $\overrightarrow{w}$
        \State $\hat{s}\gets$ replace $w_i$ with similar word $w_0$
        \State ${ss}\gets$ insert Tf-idf word in random position in $s$
        \State $\hat{ss} \gets$ merge $\hat{s}$ and ${ss}$
            
        \EndFor
        \State \textbf{return} $\hat{ss}$ \Comment{Augmented sentence}
        \EndProcedure

  \end{algorithmic}
  \caption{Word2Vec and Tf-idf based augmentation}
\end{algorithm}

\subsubsection{Masked Language Model (MLM) augmentation}
Since the above methods do not consider the context of the sentence, we decided to use Masked Language Modeling(MLM) where we used RoBERTa \cite{2019arXiv190711692L} a transformer model that is pretrained on a large corpus of English data in a self-supervised fashion. It is used to predict masked words based on the context of the sentence. You can find the algorithm used in Algorithm \ref{tab:algorithm2}. Taking the sentence, the model randomly masks 15\% of the words in the input then runs the entire masked sentence through the model and predicts the masked words which helps the model to learn a bidirectional representation of the sentence. In this work, a sentence is passed through our algorithm which then predicts the masked word creating a new augmented sentence. Note that this augmentation method was only done on the English language due to lack of enough resources to train a good MLM for the Swahili language.

\begin{algorithm}
  \label{tab:algorithm2}
\end{algorithm}

For our experiments we combined the augmented sentences for Word2Vec based-augmentation and Tf-idf based data augmentation producing almost triple the original sentences. The MLM-based augmentation methods produced almost double the original parallel sentences. The total training data that was used is as shown in Table \ref{tab:table2}.

\begin{table}[ht]
\centering
\begin{tabular}{ |p{3cm}||p{3cm}| }
 \hline
 Method \& Language Pair&Total\\
 \hline
    Word2Vec + tfidf & 2952864\\ 
    (EN-SW) &\\
    Word2Vec + tfidf & 2774186\\ 
    (SW-EN) &\\
    MLM & 2048613\\ 
    (EN-SW)&\\
 \hline
\end{tabular}
\caption{Data statistics showing total data used for training after augmentation}
\label{tab:table3}
\end{table}
%
\section{Experiments}
\label{experiments}
This section explains in detail the learning and the model settings that were considered for every data augmentation approach. 

\subsection{Model Settings}
All the models were trained using the transformer architecture of \cite{2017arXiv170603762V} using the open-source machine translation toolkit joeyNMT by \cite{kreutzer-etal-2019-joey}. The model parameters were set to 512 hidden units and embedding dimension, 4 layers of self-attentional encoder decoder with 8 heads. The byte pair encoding embedding dimension was set to 256. Adam optimizer is used throughout all experiments with a constant learning rate of 0.0003 and dropout was set at 0.3. All the models were trained on 40 epochs.

\subsection{Evaluation Metrics}
The models were evaluated using in-domain test sets. The performance of the different approaches was evaluated using different translation evaluation metrics: BLEU \cite{papineni-etal-2002-bleu}, METEOR \cite{banerjee-lavie-2005-meteor} and chrF \cite{popovic-2015-chrf}. BLEU(Bilingual Evaluation Understudy) is an automatic evaluation metric that is said to have high correlation with human judgements and is used widely as the preferred evaluation metric. METEOR(Metric for Evaluation of Translation with Explicit Ordering) is based on generalized concept of unigram matching between the machine translations and human-produced reference translations unlike BLEU and is calculated by getting the harmonic mean of precision and recall. ChrF is a character n-gram metric, which has shown very good correlations with human judgements especially when translating to morphologically rich languages. The higher the score of these metrics means that the system produces really good translations.

\section{Results and Discussion}
\label{results}
This section describes the results of the three methods; The baseline (S-NMT), the word2vec-based + tfidf (Word2Vec) augmentation and masked language model augmentation(MLM). Table \ref{tab:table4} shows the performance of the different data augmentation methods applied in machine translation. The BLEU scores for the EN $\leftrightarrow$ SW, domain specific best performing results are highlighted for each direction with the bolded scores displaying the overall best scores. We observe that in all the test domains, the models trained with the MLM-augmented data performed better than both the baseline and Word2Vec in most cases. These results are highly related to the fact that the MLM-based augmentations are based on contextual embeddings. The drop in performance in some cases can be due to the fact that the structure of the sentence is not necessarily preserved while doing word or synonym replacement thus making the translation not retain it's original meaning. We can also observe that there is a degradation of performance when translating into the low-resource language for the JW300 test data but for models tested on Tanzil, the degradation occurs mostly when translating into English.
The Tanzil training data that was used to train the model was quite low compared to JW300 data which explains the low scores for Tanzil as compared to JW300 data. The Word2Vec + Tf-idf based augmentations do not lead to significant improvements of the baseline model, however, the results show there is potential in using these methods in NMT especially the Masked Language Model for augmentation which proved to perform better than the Word2Vec+Tf-idf model


\begin{table*}[ht]
\begin{center}
\begin{tabular}{|c| c| c c c c c c|}
    \hline
    & &
    \multicolumn{2}{c}{\bfseries BLEU} &
    \multicolumn{2}{c}{\bfseries METEOR} &
    \multicolumn{2}{c|}{\bfseries ChrF} \\
    \hline
    \multicolumn{1}{|c|}{\bfseries Model} & \multicolumn{1}{c|}{\bfseries Domain} & & & & & \multicolumn{2}{c|}{\bfseries } \\
    
    && en-sw&sw-en & en-sw&sw-en & en-sw&sw-en\\
    \hline \hline
    S-NMT(BPE)& JW300 & 45.30& 46.54 & 66.32 & 62.21&65.92 & \textbf{71.30}\\
     & Tanzil & 27.48 & 24.66 & 50.43& \textbf{50.41} & \textbf{52.51}&  46.69\\ 
    \hline
    Word2Vec & JW300 &  45.23& 45.52& 65.90& 66.56& 65.02&61.76\\
                & Tanzil &  26.29&  \textbf{25.80}& 49.45& 42.43& 45.78& 40.68\\
    \hline
     MLM & JW300 &  \textbf{45.32}&  \textbf{46.98}& \textbf{66.56}& \textbf{70.55}& \textbf{65.94}&69.68\\
        & Tanzil &  \textbf{29.26}& 24.86 & \textbf{58.31}& 49.31& 48.23&\textbf{47.47}\\
    \hline
    
\end{tabular}
\caption{BLEU, ChrF, Meteor scores for $Swahili \leftrightarrow English$ directions, domain-specific best performing results are in bold.}
\label{tab:table4}
\end{center}
\end{table*}






\section{Limitations and Future Work}
\label{limitations}
One of the biggest challenges in machine translation today is learning to translate low-resource language pairs with technical challenges such as learning with limited data or dealing with languages that are distant from each other. 

This paper shows that we could potentially use simple data augmentation methods in machine translation. In our experiments, we only augment the source language for the Word2Vec based augmentation method and only augment the English sentences for the MLM based augmentations. In Future work, we plan on exploring augmenting the target side of the parallel data in the Word2Vec-based augmentation and compare performance to the source language augmentation as well as testing the model's ability to generalize by using out-of domain datasets. Another experiment that could be explored is the use of the Word2Vec data augmentation method only without the use of Tf-idf word replacement method as it adds more noise to the sentences. We plan on continuing this research and will make available the algorithms used in this paper at \textit{\url{https://github.com/dsfsi/translate-augmentation}}

\subsubsection{Computational Considerations}
Training time took about 1 hour running one epoch using NVIDIA Tesla V100 GPU on Google Cloud on the augmented texts. Running on Colaboratory took about 5 days to run 40 epochs and with limited time on our hands, there is only so much we could experiment. Running these experiments was quite expensive and there needs to be consideration of budgets as well as time so as to run these MT experiments.   

\section{Conclusion}
\label{conclusion}
In this work we proposed the use of different textual data augmentation tasks in neural machine translation using the low-resourced language Swahili. We also showed how one can perform data augmentation on the low resourced language using pre-trained word vectors and presented baseline results in ChrF and METEOR which have never been presented before. Our investigation shows that although the models trained on the augmented texts did not improve on the baseline model, there is still potential to using these methods in NMT tasks with enough compute and more experiments. We hope that this work will set the stage for further research on applying simple augmentation methods that don't require a lot of computation power in low-resource NMT modelling.

\section{Bibliographical References}\label{reference}
\bibliographystyle{agsm}
\bibliography{bibliography.bib}
\end{document}